\documentclass[conference]{IEEEtran}
\makeatletter
\def\ps@headings{%
\def\@oddhead{\mbox{}\scriptsize\rightmark \hfil \thepage}%
\def\@evenhead{\scriptsize\thepage \hfil \leftmark\mbox{}}%
\def\@oddfoot{}%
\def\@evenfoot{}}
\makeatother
\usepackage[justification=centering]{caption}
\usepackage{url}
\usepackage{booktabs}
\usepackage{graphicx,subfigure}
\usepackage{epstopdf}
\usepackage{amsmath}
\usepackage{algorithm}
\usepackage{algpseudocode}
\usepackage{amsmath}
\usepackage{amssymb}
\usepackage{amsthm}
\usepackage{epsfig}

\usepackage{amsfonts}
\usepackage{multirow}
\usepackage{color}
\usepackage{array}
\usepackage{listings}
\usepackage{hyperref}
\usepackage[underline=true]{pgf-umlsd}

\usepackage{setspace}

\begin{document}

\title{Voice Recognition Robot with Real-Time Surveillance and Automation}

\author{\IEEEauthorblockN{Lochan Basyal}
\IEEEauthorblockA{\textit{Stevens Institute of Technology, Hoboken, NJ, USA} \\
bashyallochan@gmail.com, lbasyal@stevens.edu}
}

\maketitle

\begin{abstract}
Voice recognition technology enables the execution of real-world operations through a single voice command. This paper introduces a voice recognition system that involves converting input voice signals into corresponding text using an Android application. The text messages are then transmitted through Bluetooth connectivity, serving as a communication platform. Simultaneously, a controller circuit, equipped with a Bluetooth module, receives the text signal and, following a coding mechanism, executes real-world operations. The paper extends the application of voice recognition to real-time surveillance and automation, incorporating obstacle detection and avoidance mechanisms, as well as control over lighting and horn functions through predefined voice commands. The proposed technique not only serves as an assistive tool for individuals with disabilities but also finds utility in industrial automation, enabling robots to perform specific tasks with precision.

\end{abstract}


\section{Introduction}
A robot is a complex machine incorporating electrical, mechanical, communication, and computing components to effectively perform specific tasks. Robotics, as a technology, aims to minimize the necessity for human involvement in various work processes. While speech recognition involves the system comprehending spoken words, it does not necessarily grasp their meaning\cite{lbasyal}. Voice recognition robots, on the other hand, operate based on predefined voice commands. The initial step involves processing a voice command through the Android platform, where the conversion of voice to text occurs within the system. The development of this Android application utilizes a drag-and-drop programming technique and is implemented with the MIT App Inventor. This platform is chosen for its simplicity compared to other development platforms, and the Google speech recognizer is employed for an effective system response. Establishing a communication network necessitates Bluetooth and internet connectivity.
The robot's control mechanism relies on various commands such as right, left, backward, forward, light on, horn please, etc. These commands are processed through a smartphone, and the real-world operation is accomplished through a control circuit interfacing with the motor mechanism. The Bluetooth technology utilized for automation is cost-effective compared to alternative communication platforms like Zigbee, GSM (Global System for Mobile Communication), and General Packet Radio Service (GPRS), with a simple installation process. Research indicates that Bluetooth systems are faster than GSM systems\cite{lbasyal}. Furthermore, the integration of a motor driver and relay driver module with Arduino enhances the effectiveness of driving output loads. This paper also emphasizes real-time surveillance and automation concepts, incorporating an obstacle detection and avoidance mechanism facilitated by the ultrasonic sensor HCSR04. The ultrasonic sensor measures the distance between the target and the robot's actual position. This is achieved by transmitting an ultrasonic wave that reflects off a target, and the time taken for the echo to return helps calculate the distance. The robot's movement is determined based on the distance measurement by the ultrasonic sensor, and this concept is implemented through programming in the Arduino IDE (Integrated Development Environment) and loading it onto the Arduino Atmega 328.
The subsequent sections of this paper are organized to provide a comprehensive understanding of the development of this technology. Section II delves into the crucial technical components necessary for its implementation. In Section III, the simulation of this concept has been executed using Proteus simulation software, and the results are presented. The programming concept of the robot is elucidated in Section IV, detailing the software aspects essential for its functionality. Moving on to Section V, the design architecture of this technology is thoroughly depicted, offering insights into the overall structural framework. Similarly, Section VI is dedicated to the hardware implementation of this concept. It provides a detailed account of how the theoretical concepts are translated into a physical system, including the integration of components such as motor drivers, relay modules, and Arduino boards. Finally, the paper is concluded in Section VII, summarizing the key findings and implications of the study. Additionally, possible avenues for future work in the field are explored.

\section{Technical Components}

\subsection{Arduino}
\begin{figure}
    \centering
    \includegraphics[width= 0.7\linewidth]{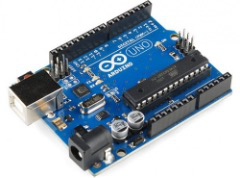}
    \caption{Arduino Uno ATmega328\cite{lbasyal}}
    \label{arduino}
\end{figure}

The first essential technical component for this project is the Arduino Uno (Figure\ref{arduino}), a microcontroller based on the Atmega328\cite{atmega328p}. It encompasses 14 digital input/output pins, with 6 of them designated as pulse width modulation (PWM) outputs. Additionally, it features 6 analog pins designed for sensor interfacing. The inclusion of a 16 MHz ceramic resonator\cite{lbasyal} and the incorporation of all necessary circuitry within a single module contribute to the operational simplicity of the Arduino Uno. This microcontroller is chosen for its user-friendly design and ease of operation compared to other microcontroller circuits.

\subsection{Bluetooth Module HC-05}
\begin{figure}[h]
    \centering
    \includegraphics[width=0.5\linewidth]{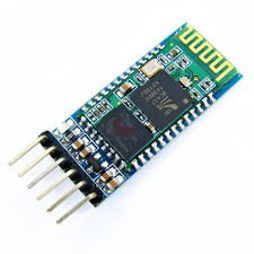}
    \caption{Bluetooth module HC-05\cite{lbasyal}}
    \label{bluetooth}
\end{figure}

The Bluetooth module (HC-05) (Figure\ref{bluetooth}) functions as an interface with an Arduino circuit to establish a communication platform. Primarily, this module is utilized in this project for the reception of serial data from an Android application. In practice, the module operates within a range of 50 meters\cite{lbasyal}.

\subsection{PCB Design of L293D Module}
\begin{figure}[h]
    \centering
    \includegraphics{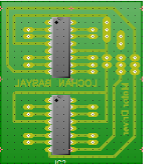}
    \caption{PCB design real-world view\cite{lbasyal}}
    \label{pcbdesign}
\end{figure}

\begin{figure}[h]
    \centering
    \includegraphics{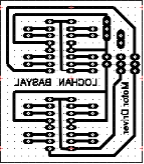}
    \caption{PCB design artwork view\cite{lbasyal}}
    \label{pcbdesign1}
\end{figure}

The L293D is an H-bridge motor driver IC with an operating voltage range of up to 36V from 5V. It provides a bidirectional drive current of up to 600mA, enabling control over two DC motors and their directions. Speed control is achieved using pulse width modulation (PWM)\cite{lbasyal}. For this project, a motor driver module was created through PCB (Printed Circuit Board) design, with each module containing two L293D ICs. Figure\ref{pcbdesign} shows the real-world view of designing on PCB Wizard, whereas Figure\ref{pcbdesign1} represents the design artwork view.

\subsection{Control Circuit Module}
\begin{figure}[h]
    \centering
    \includegraphics[width=0.7\linewidth]{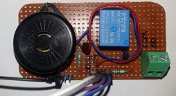}
    \caption{Bluetooth module HC-05\cite{lbasyal}}
    \label{controlcircuit}
\end{figure}

This module incorporates a relay driver mechanism with the BC547 transistor and a buzzer, serving as the robot's horn. It activates any output system through a relay by applying a digital pulse through an Arduino. This segment of the project will govern the lighting and horn mechanisms of the robot. The design of this module is depicted in Figure\ref{controlcircuit}.

\subsection{Ultrasonic Sensor HCSR04}
\begin{figure}[h]
    \centering{\includegraphics[width=0.7\linewidth]{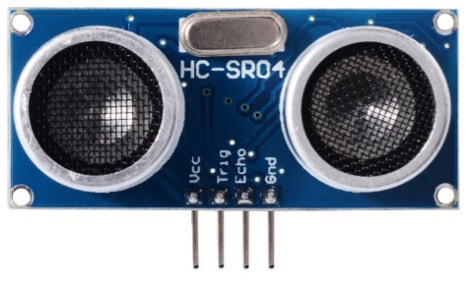}}
    \caption{Ultrasonic sensor HCSR04\cite{lbasyal}}
    \label{ultrasonic}
\end{figure}

The interfacing mechanism between an Arduino and the ultrasonic sensor HCSR04 involves measuring the distance between a target and the real position of the sensor. This is achieved by calculating the time duration between the transmission and reception of an ultrasonic wave. The module contains four terminals: Vcc, Trigger, Echo, and Gnd. Figure\ref{ultrasonic} illustrates the module.

\subsection{Android Application}
\begin{figure}[h]
    \centering{\includegraphics[width=0.8\linewidth]{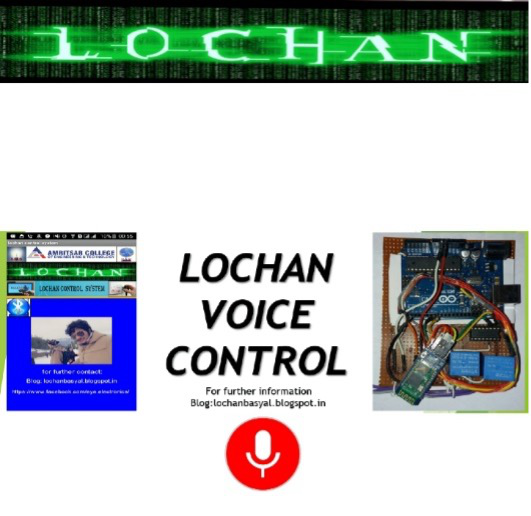}}
    \caption{Android Application\cite{lbasyal}}
    \label{androidapp}
\end{figure}

The Android application (Figure\ref{androidapp}) serves as the primary interface for voice signals and has been developed using the MIT App Inventor\cite{mitappinventor} platform, employing a drag-and-drop programming approach. The application operates based on Bluetooth and internet connectivity and was designed with the utilization of the Google Voice Recognizer module. Its functionality includes converting voice commands into corresponding text messages displayed on the screen. Subsequently, this message is transmitted to an electronic hardware circuit through Bluetooth connectivity. Upon receiving the text generated from the voice command, the system employs an 'if' statement to check whether the command matches a predefined set of commands. If a match is detected, the system activates, initiating the execution of the programmed actions. In practical terms, this activation results in the Arduino providing a digital high signal, delivering 5 volts of direct current (DC). This approach allows for a responsive and conditional control mechanism where the system's behavior is contingent on the recognized voice command meeting specific predefined criteria.

\section{Proteus Simulation Software}
\begin{figure*}[ht!]
    \centering
    \includegraphics[width=0.8\textwidth]{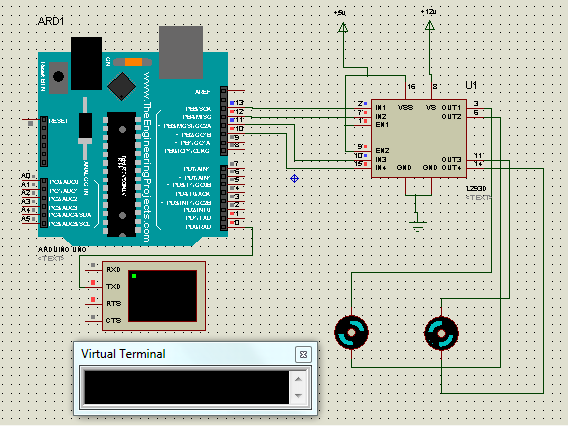}
    \caption{Simulation of the motor mechanism using Proteus\cite{lbasyal}.}
    \label{simulation}
\end{figure*}

Proteus simulation software, version 7, is utilized to validate circuit connections in the virtual realm. Using this platform, a circuit can be drawn, and the programming file in hexadecimal format—generated after compiling the program in the Arduino IDE—can be uploaded. In this simulation, Arduino interfaces with the L293D IC as the motor driver. The communication platform is established through the virtual terminal, and a circuit has been identified that operates according to a programmed sequence. The simulation of the concept is presented in Figure\ref{simulation}.

\section{Programming Concept of the Robot}
\begin{figure}[h]
    \centering
    \includegraphics{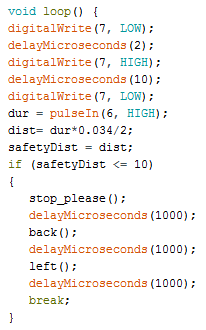}
    \caption{Ultrasonic sensor interfacing\cite{lbasyal}}
    \label{code1}
\end{figure}

\begin{figure}[h]
    \centering
    \includegraphics{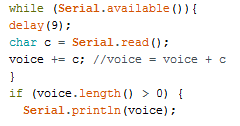}
    \caption{Code for string processing\cite{lbasyal}}
    \label{code2}
\end{figure}

\begin{figure}[h]
    \centering
    \includegraphics{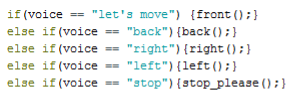}
    \caption{Code for voice command execution\cite{lbasyal}}
    \label{code3}
\end{figure}

\begin{figure}[h]
    \centering
    \includegraphics{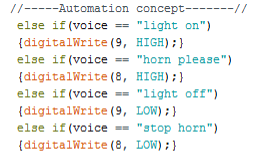}
    \caption{Code for automation concept\cite{lbasyal}}
    \label{code4}
\end{figure}

The communication platform established in our project is Bluetooth, based on serial data transmission and reception. When serial data is available, the program reads the bytes from Bluetooth, initiates string processing, and checks the received string against predefined voice commands. Upon a match, the program executes a statement. Function definitions illustrating the robot's movement mechanism have been presented in this paper. 
Figure\ref{code1} illustrates the code for ultrasonic sensor interfacing with an Arduino, while Figure\ref{code2} depicts the code for string processing. Similarly, Figure\ref{code3} showcases the code for voice command execution, and Figure\ref{code4} introduces the concept of automation, demonstrating the control of lighting and horn mechanisms through voice commands.
In this program, the voice is defined as a string, and parameters such as distance, duration, and safety distance are declared as integer, long, and integer, respectively. These parameters are represented by their short forms, as shown in Figure 10. The ultrasonic sensor is employed for obstacle detection and avoidance, a crucial condition applied to the robot. This sensor has four pins for Vcc (DC 5V supply voltage), Ground (GND), trigger pin, and echo pin.
Distance measurement is facilitated by the pulse received from digital pin 6 of an Arduino, utilized as the echo pin in our project. Similarly, digital pin 7 represents the trigger pin. Upon detecting an obstacle, the robot halts immediately, moves backward, and takes a left direction, as per the code depicted in Figure\ref{code1}. The robot ceases its movement within 10 centimeters of obstacles, and after completing subsequent movements, the safety distance condition is removed from the programming loop.
In real-world operation, when the user provides a voice command through an Android application and the string check condition becomes true, the program executes based on the function call within that particular condition. This process demonstrates real-world operation, meaning the robot's movement in a specified direction. Real-time distance surveillance is conducted by the ultrasonic sensor.

\section{Design Architecture}
\begin{figure}[h]
    \centering
    \includegraphics{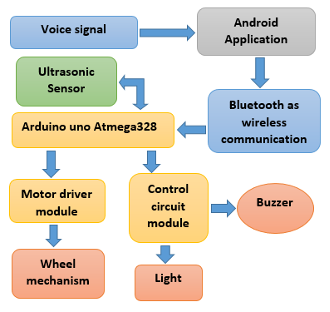}
    \caption{Design Architecture\cite{lbasyal}}
    \label{design}
\end{figure}

\begin{figure}[h]
    \centering
    \includegraphics{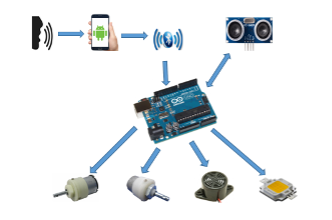}
    \caption{Block Diagram\cite{lbasyal}}
    \label{design1}
\end{figure}

Design Architecture illustrates the overall mechanism of the proposed methodology. Figures\ref{design} (design architecture) and\ref{design1} (block diagram) depict the different sections of the project, encompassing two main components: an Android application and a robot equipped with a wheel mechanism, light, and buzzer. In this methodology, wireless communication is established through serial communication, where one bit at a time is transmitted and can be processed by a controller. This setup forms the foundation for implementing the proposed system.

\section{Hardware Implementation}
\begin{figure}[h]
    \centering
    \includegraphics[width=\linewidth]{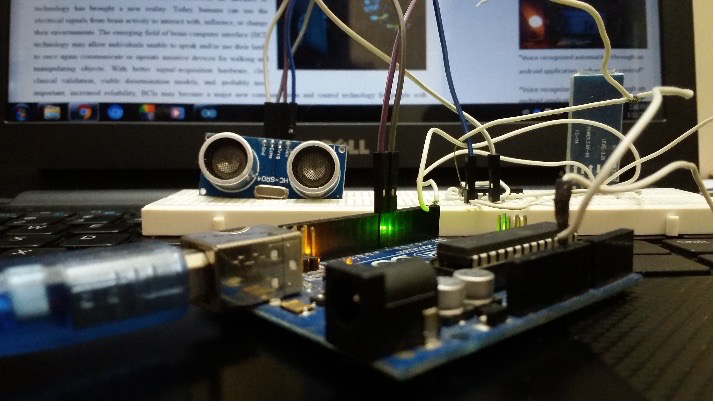}
    \caption{Hardware Testing\cite{lbasyal}}
    \label{htesting}
\end{figure}

\begin{figure}[h]
    \centering
    \includegraphics[width=\linewidth]{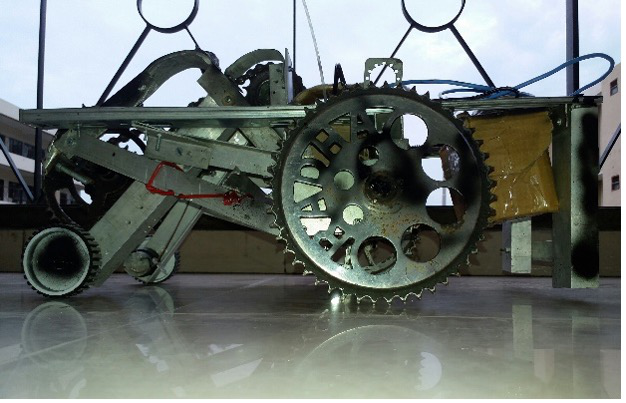}
    \caption{Hardware Implementation\cite{lbasyal}}
    \label{himplement}
\end{figure}

This section provides a detailed account of how the theoretical concepts are translated into a physical system, encompassing the integration of components like motor drivers, relay modules, and Arduino boards.
In the hardware testing phase, rigorous evaluation and validation were conducted to ensure the seamless functioning of each component. This involved testing the responsiveness of the motor drivers, assessing the reliability of the relay modules, and confirming the proper integration of Arduino boards within the system. The objective of this phase was to identify and address any potential issues before proceeding to the full-scale hardware implementation.
Figure\ref{htesting} illustrates the hardware testing process, showcasing the systematic evaluation of individual components and their collective functionality. This phase aimed to verify the robustness of the hardware setup, ensuring that each element performs as intended and is ready for the subsequent stages of development.
Moving forward, the hardware implementation phase involved the physical development of the robot, integrating the tested components into a cohesive and functional system. The construction included mounting the motor drivers, relay modules, and Arduino boards onto the robot chassis, aligning with the design architecture presented in earlier sections.
Figure\ref{himplement} visually represents the completed hardware implementation, illustrating the integrated components within the physical structure of the robot. The design carefully incorporates voice recognition technology, highlighting how the hardware components seamlessly interact with the software system.

The integration of voice recognition technology into the physical robot marks a crucial advancement, bringing the concept to life. This section emphasizes the comprehensive process of hardware testing, ensuring the reliability of individual components, followed by the successful physical development of the robot. The figures provide a visual representation of both the testing and implementation phases, offering a complete understanding of the hardware journey in the realization of the voice recognition robot concept.

\section{Conclusion with Future Work}
This paper successfully demonstrates the concept of a voice recognition robot with real-time surveillance and automation, integrating both software and hardware implementations. The potential use cases of this concept have the potential to significantly enhance the ease of human life by automating control over manual operations. The concept presented in this paper could be further improved in terms of the recognition system's performance and the matching of predefined commands with user input commands. This improvement involves calculating the similarity between two different vector embeddings representing user input and the predefined command.
Implementing this enhancement would allow flexibility in user input, meaning the command provided by the user wouldn't need to precisely match the predefined command for the robot to execute the desired action. This approach addresses issues encountered when the robot is handled by multiple users, ensuring more robust and user-friendly interaction. Future work in this area could focus on refining the recognition system, exploring additional use cases, and advancing human-robot interaction for broader applications in various scenarios.

\section*{Acknowledgment}
   The author would like to acknowledge the foundation laid by his previous work\cite{lbasyal}. This paper represents an iteration aiming to enhance clarity and eliminate typos.

\bibliographystyle{IEEEtran}
% \bibliography{} 

\end{document}